\pdfoutput=1

\documentclass[11pt]{article}

\usepackage[preprint]{coling}
\usepackage{graphicx}
\usepackage{enumitem}
\setlist[itemize]{noitemsep, topsep=0pt}

\usepackage{times}
\usepackage{latexsym}
\usepackage{geometry}
\usepackage{mathastext}
\usepackage{float}
\usepackage{array}
\usepackage{multirow}
\usepackage{tabu}
\usepackage{listings}
\usepackage{amsmath}

\usepackage{makecell}
\usepackage{tabularx}
\usepackage{geometry}
\usepackage{longtable}

\usepackage[T1]{fontenc}

\usepackage[utf8]{inputenc}

\usepackage{microtype}
\usepackage{xcolor}
\usepackage[safe]{tipa}
\usepackage{multirow}
\usepackage[T1]{fontenc}
\usepackage{wrapfig}
\usepackage{adjustbox}
\usepackage{tabularx}
\usepackage{adjustbox}
\usepackage{colortbl}
\usepackage{xcolor}

\usepackage{inconsolata}

\title{Transforming NLU with Babylon: A Case Study in Development of Real-time, Edge-Efficient, Multi-Intent Translation System for Automated Drive-Thru Ordering}

\author{
    Mostafa Varzaneh*, Pooja Voladoddi*, Tanmay Bakshi*, Uma Gunturi* \\
    IBM watsonx Orders \\ 
}

\begin{document}
\maketitle
\begin{abstract}
Real-time conversational AI agents face challenges in performing Natural Language Understanding (NLU) in dynamic, outdoor environments like automated drive-thru systems. These settings require NLU models to handle background noise, diverse accents, and multi-intent queries while operating under strict latency and memory constraints on edge devices. Additionally, robustness to errors from upstream Automatic Speech Recognition (ASR) is crucial, as ASR outputs in these environments are often noisy. We introduce Babylon, a transformer-based architecture that tackles NLU as an intent translation task, converting natural language inputs into sequences of regular language units (‘transcodes’) that encode both intents and slot information. This formulation allows Babylon to manage multi-intent scenarios in a single dialogue turn. Furthermore, Babylon incorporates an LSTM-based token pooling mechanism to preprocess phoneme sequences, reducing input length and optimizing for low-latency, low-memory edge deployment. This also helps mitigate inaccuracies in ASR outputs, enhancing system robustness. While this work focuses on drive-thru ordering, Babylon’s design extends to similar noise-prone scenarios, for e.g. ticketing kiosks. Our experiments show that Babylon achieves significantly better accuracy-latency-memory footprint trade-offs over typically employed NMT models like Flan-T5 and BART, demonstrating its effectiveness for real-time NLU in edge deployment settings. 
\end{abstract}

\renewcommand{\thefootnote}{\fnsymbol{footnote}}
\footnotetext[1]{These authors contributed equally to this work.}
\footnotetext[2]{Emails ids of the corresponding authors are: uma.gunturi@ibm.com, pooja.voladoddi@ibm.com.}

\section{Introduction} \label{introduction}
\begin{figure}[ht]
\centering
\includegraphics[width=1.0\columnwidth]{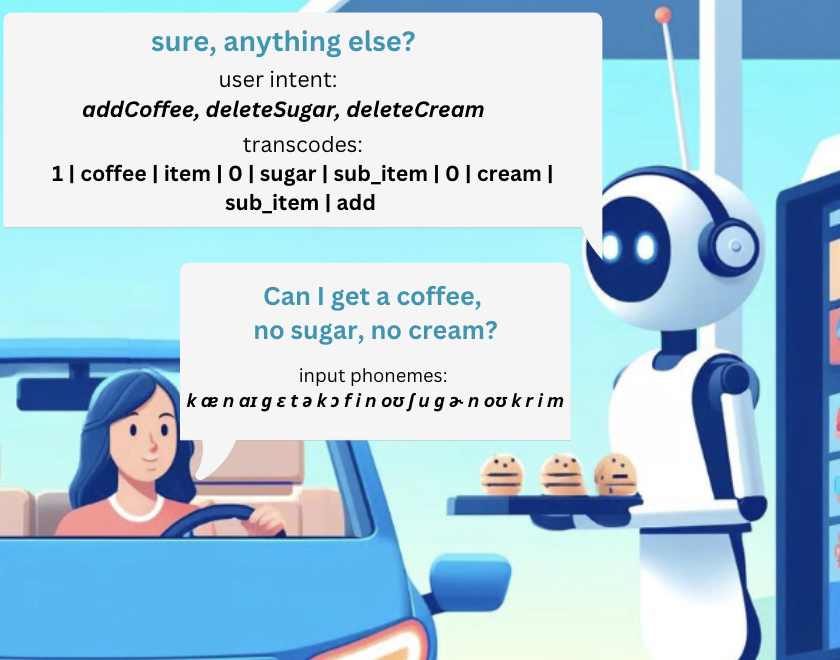}
\caption{The input to our NLU model for this customer order is a sequence of \textbf{phonemes}. The output generated by the NLU model post translation \texttt{[1 | coffee | item 0 | sugar| sub\_item | 0 | cream | sub\_item | add]} as seen in the figure, is a sequence of output \textbf{transcodes} and represents the customer intent to add a \textit{coffee} to their order and remove \textit{sugar} and \textit{cream} from it.}
\label{fig:workflow-fig}
\end{figure}
In the evolving landscape of conversational AI, efficient natural language understanding components are crucial, particularly for real-time, edge-deployed applications like automated drive-thru ordering. A recent trend has framed intent classification and slot filling as a translation task \cite{de-bruyn-etal-2022-machine, 10180229, hoscilowicz2024large}, transforming user utterances into structured outputs with intent and slot labels. However, not many works have captured effort entailed in making this approach highly performant and efficient for low-latency, edge-deployed systems. To address this, we propose a method that represents intents and slots as regular language units. This representation enhances processing efficiency, particularly in edge environments with limited computational resources and memory. Furthermore, we introduce \textit{Babylon}, a custom model optimized for edge deployment, which uses LSTM-based token pooling alongside a Transformer architecture to manage multi-intent scenarios efficiently. This enhancement plays a critical role in reducing the computational load of the Transformer and mitigating errors from upstream Automatic Speech Recognition (ASR) components. Unlike other methods that rely on complex mechanisms like chunked or windowed attention to enhance Transformer performance \cite{ ju2021chunkformer, Beltagy2020LongformerTL}, Babylon maintains simplicity, minimizing computational overhead—a key advantage for latency-sensitive applications such as drive-thru systems.

Additionally, the representation of intents and slots as regular language units allows Babylon to efficiently handle multi-intent scenarios while maintaining real-time edge performance offering a smooth user experience. For example, refer to Fig. \ref{fig:workflow-fig} where customer in their turn of the dialogue asks for a coffee, and simultaneously to modify their coffee to have no sugar, no cream. These optimizations are vital for edge-deployed environments, where minimizing latency, memory usage, and error propagation is critical. The key contributions of this paper include:

(1) We propose an intent translation framework for real-time NLU that uses regular language units called \textit{"transcodes"} to represent actionable customer intent, enhancing efficiency in resource-constrained edge deployments. This representation is particularly suitable for the dynamic and ambiguous nature of spoken queries encountered in drive-thru scenarios.

(2) The Babylon model integrates LSTM-based token pooling with a Transformer backbone to process multiple user queries in a single dialogue turn. This architecture avoids the computational complexity of chunked or windowed attention and simplifies deployment on edge devices, achieving significant improvements in latency and memory usage.

(3) Our experimental results show that Babylon outperforms other NMT-based architectures, including models like Flan-T5 and BART, in terms of accuracy-latency trade-offs for edge environments. Babylon’s capability to operate on a single CPU core in a single-threaded mode demonstrates its practical utility for real-world deployment.

By building on existing translation-based methods for intent detection and slot filling and advancing the approach through the use of regular language units, this case study offers valuable insights for developing efficient NLU systems for edge-deployed conversational AI agents.

\section{System Description}\label{problem}
\begin{figure}[ht]

\centering
\includegraphics[width=0.48\textwidth]{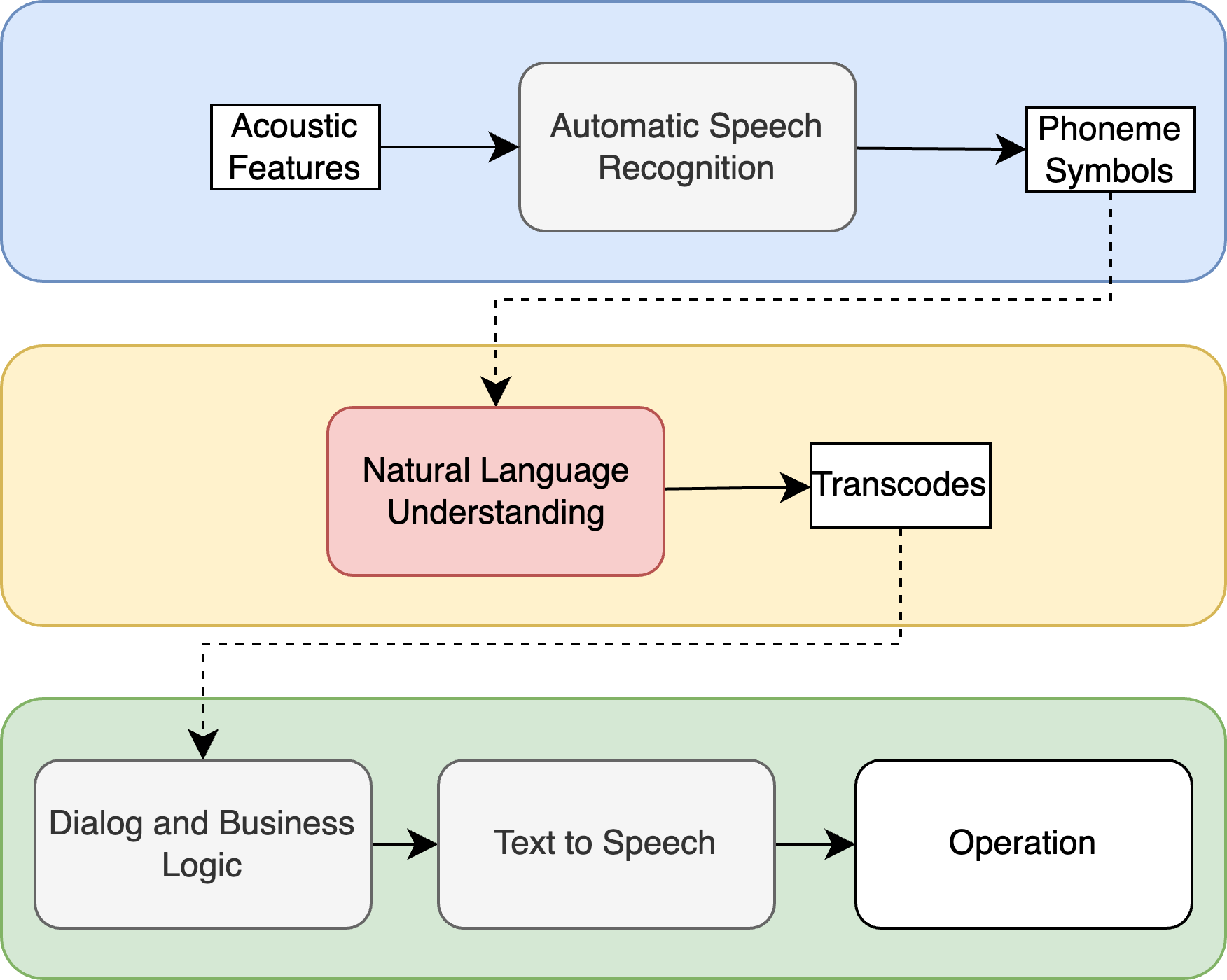}
\caption{Illustration of our overall system. We re-purpose the NLU task of intent classification and slot filling into a translation task. 
We translate from audio units (\textit{phonemes}) to intent language units (\textit{transcodes}).}
\label{fig:overall_asr_workflow} 
\end{figure}

As depicted in Figure \ref{fig:overall_asr_workflow}, our drive-thru conversational AI integrates four core components: Automatic Speech Recognition (ASR), Natural Language Understanding (NLU), Dialog Management, and Text-To-Speech (TTS) synthesis. This setup enables real-time, accurate customer service interactions with limited compute resources. The workflow is as follows (Refer to Figure \ref{fig:overall_asr_workflow}):

\begin{enumerate}[noitemsep, topsep=0pt, partopsep=0pt]
\item Neural ASR transcribes real-time audio into \textit{phonemes}, input for the NLU component.
\item NLU converts phonemes into regular language units, \textit{transcodes}, representing customer intents.
\item Dialog Management processes transcodes, \textit{updates} order state, and triggers TTS to respond to the customer.
\end{enumerate}
\vspace{5pt}


\vspace{-5pt}
This work focuses specifically on the \textbf{NLU component} that employs a NMT model, whose input is a sequence of phonemes, and output is a sequence of \textit{"transcodes"}. For e.g, given a natural language drive-thru order (speech) \textbf{p} that can be transcribed into a \textit{phoneme sequence} \textbf{p} $ = [\mathit{p}_{1}, \mathit{p}_{2},..., \mathit{p}_{i},..., \mathit{p}_{m}]$, goal is to predict the customer intent, a \textit{transcode sequence} \textbf{t} $\mathit{ = [t_{1}, t_{2}, t_{3}, ..., t_{i},... t_{n}]}$. Here, $\mathit{p_i}$ is a phoneme; unit of textual representation of sound emitted from our ASR component, and transcode $\mathit{t_{i}}$ is a token (word) of an internally created regular language that represents complex customer intents. Note that "complex" in this context means compositionally representing multiple intents and slots referenced within one inference call. For e.g. \textit{"Hi, can i get two chocolate chip cookies please ... ah ... actually no ... can you just do two peanut butter cookies instead?"} - this single turn within a customer order contains multiple intents such as two \texttt{addItem} intents and one \texttt{deleteItem} intent in reference to three different entities (quantity, item names), all processed within a single customer turn (a single inference call.)
 
 

\setlength{\intextsep}{3pt plus 1pt minus 1pt} 

 \section{Related Work} \label{related_work}


\subsection{Conventional NLU approaches}

Task-oriented dialogue systems have traditionally employed \textbf{(1)} separate models for intent detection (ID) and slot filling (SF), such as support vector machines, conditional random fields \cite{xu2014targeted} and recurrent neural networks (RNNs) of various types \cite{kurata-etal-2016-leveraging}, which often result in error propagation and suboptimal performance due to underutilized cross-task dependencies \cite{10.5555/2817174.2817185, 9533525, Qin2021ASO}; \textbf{(2)} joint models with separate decoders or layers for ID and SF \cite{liu2016attention, 6707709, guo2014joint, hakkani-tr2016multi-domain}, which often face issues such as increased computational complexity, higher latency, and greater memory usage, making them unsuitable for resource-constrained environments \cite{weld2022survey, 10232869}. \textbf{(3)} Semantic parsing demands a large amount of annotated data for mapping each utterance to its semantic representation and faces challenges with the dynamic nature of spoken language \cite{aghajanyan-etal-2020-conversational}.

Recently, there has been a paradigm shift towards transformer-based sequence-to-sequence frameworks for NLU tasks, including Neural Machine Translation (NMT), Question Answering, and Semantic Parsing \cite{gardner-etal-2018-neural, Raffel2019ExploringTL, weld2022survey, de-bruyn-etal-2022-machine, fitzgerald-2020-stil, luong2015effective}. 
However, these methods use well-known Transformers such as XLM-Roberta \cite{conneau-etal-2020-unsupervised}, mT5 \cite{xue-etal-2021-mt5} and mBART \cite{liu2020multilingual}. NMT models like this are too large to be efficiently deployed on edge CPU devices without compromising performance. Given these limitations, there is a growing interest in developing edge-efficient NMT models that strike a balance between performance and resource usage.

\subsection{Neural Machine Translation for Multi-Intent Detection}
NMT has proven highly effective in various domains, transforming input sequences into output sequences for structured tasks like regex modeling \cite{locascio-etal-2016-neural} and formal languages such as SQL \cite{9138181, bandyopadhyay-zhao-2020-natural}. While works such as \cite{de-bruyn-etal-2022-machine, 10180229, hoscilowicz2024large} have explored NMT-based approaches for ID and SF, to the best of our knowledge, their application to low latency and edge-efficient multi-intent detection remains largely unexplored.

To this end, this paper proposes \textit{Babylon}, a custom NMT-based architecture designed for translating user queries represented in the form of "\textit{phonemes}," into regular language units referred to as "\textit{transcodes}," which encapsulate complex customer intents. \textit{Babylon} is tailored for multi-intent detection in a real time drive-thru order-taking scenario, operating on resource-constrained edge devices without cloud connectivity.

\subsection{Limitations of vanilla Transformers} \label{limitations_transformer}

\textbf{Computational Complexity:} The standard Transformer's time and memory complexity, scales with input size and layers, and is impractical for resource-constrained settings, especially with long input sequences. Sub-word tokens like phoneme-based textual representation further increases input length, worsening computational demands. For e.g.,\textit{"can I get a coffee please?"} becomes \texttt{k \textreve \hspace{3pt}n a\textsci\hspace{2pt} \textg\hspace{2pt} \textsci\hspace{2pt} t\hspace{2pt} \textreve\hspace{2pt} k \textopeno\hspace{2pt} f i p l i z} which has 16 tokens rather than 6.
     
\noindent\textbf{Loss of Sequential Information:} While Transformers can handle long sequences, they fundamentally rely on positional encodings \cite{chen2021simple} and lack an inherent mechanism to capture sequential order \cite{haviv2022transformer}. This can be suboptimal for tasks where the order of input elements is critical, such as in our drive-thru use case. For example, people tend to repeat themselves when ordering food (\textit{"cookie uh cookie ... chocolate"}), produce false starts (\textit{"and for the cookie... uh yea the chocolate cookie"}), or repeat entire phrases (\textit{"I'll get the cookie... uh yeah I'll just get a cookie"}). Here, the challenge for the model is to detect and correct duplicated intents based on the position, distance and context.
   
\noindent\textbf{Inadequate Local Context Representation}: Transformers primarily focus on global dependencies \cite{huang2023advancing}, often failing to capture the local context needed for phoneme-based inputs. Phoneme embeddings also carry less contextual information than word or subword embeddings \cite{Sharma2021PhoneticWE}, making them less effective at conveying user intent or subtle meaning differences \cite{huang2023advancing, Sharma2021PhoneticWE}. Previous work has combined convolutional neural networks (CNNs) and recurrent RNNs with Transformers to enhance local dependency capture and preserve temporal dynamics \cite{wu2020lite, dumitru2024enhancing}. Token pooling mechanisms within Transformers reduces sequence length and computational load, especially for computer vision applications \cite{marin2021token}. In this paper, we propose combining an LSTM with token pooling without modifying the standard Transformer architecture to address phoneme-based input challenges.

\section{Babylon Architecture} \label{babylon_architecture}

\begin{figure}[ht]

\centering
\includegraphics[width=0.5\textwidth]{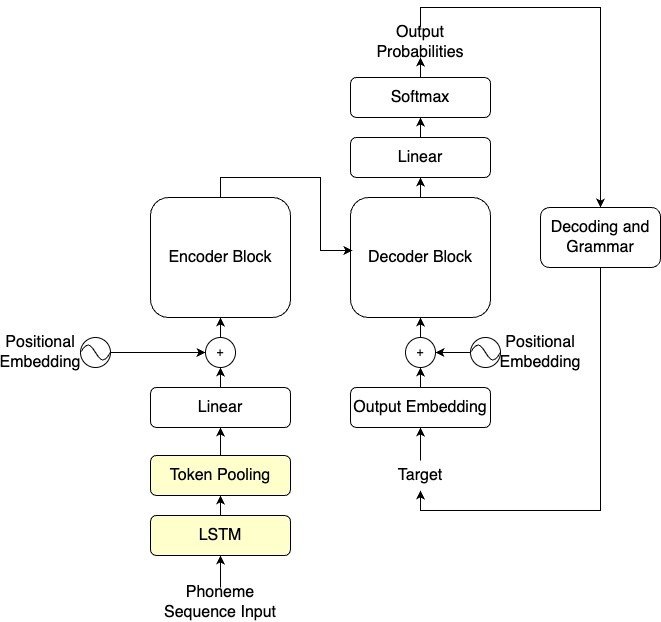}
\caption{Architecture of Babylon. Blocks highlighted in yellow represent the LSTM and Token pooling layers added prior to the original \cite{vaswani2017attention} transformer architecture. }
\label{fig:architecture-fig} 
\end{figure}

The components that are unique to Babylon are highlighted in Figure \ref{fig:architecture-fig}.
We process inputs through bidirectional LSTM network, followed by a token-pooling layer, finally passing the now context-rich, downsampled inputs to the Transformer network. We believe that this approach is easy to implement and addresses challenges discussed earlier in section \ref{limitations_transformer}:

\subsection{LSTM, the hero!}
To this end, we incorporate an LSTM layer followed by a token pooling layer prior to generating the input embedding and positional encoding.  Additionally:

\paragraph{LSTM as neural rolling memory:} The LSTM acts as a learnable memory for the transformer layer, providing context access.

\paragraph{LSTM for keeping temporal information:} Although positional embeddings capture relative token positions, transformers struggle with temporal or spatial information \cite{dosovitskiy2021image, dai2019transformerxl, zeng2022transformers} while an LSTM layer by nature preserves sequential information.

\paragraph{LSTM as regularizer:} Our findings indicate that vanilla Transformers exhibit sensitivity to noisy tokens from upstream ASR errors. By employing an LSTM with a low-dimensional hidden space, we can effectively regularize and compress information, thereby reducing sensitivity to individual token (in our case, phoneme) variations.

\subsection{Downsampling by token pooling}
Passing LSTM outputs to the transformer increases computational and memory demands. We apply token pooling, passing every $\mathit{k^{th}}$ LSTM output unit to the transformer, reducing computations and further aiding in regularization of erroneous ASR tokens. To ensure the model sees the \textit{<EOS>} token for decoding, pooling starts from the end.
In this work, we experimented with a range of $\mathit{k}$ and selected $\mathit{k}$=4 for optimal performance. 

\begin{table*}[htbp]
\centering
\resizebox{0.8\textwidth}{!}{
\begin{tabular}{|l|l|l|l|}
\hline
\textbf{Model} & \textbf{Model Type} & \textbf{\#Params} & \textbf{Training Time (per run)} \\ \hline
Atlantis & Bi-LSTM & 12.5 million & 1.2-1.5 hours \\ \hline
Delphi & Bi-LSTM + Attention & 9 million & 50 minutes-1 hour \\ \hline
Camelot & Vanilla Transformer & 5 million & 30-40 minutes \\ \hline
Babylon & Bi-LSTM Pooled Transformer & 7 million & 40-50 minutes \\ \hline
\multicolumn{4}{|l|}{\textit{\textbf{LLM based NMT models}}} \\ \hline
Flan-T5-small & Instruction-tuned & 60 million & 2-3 hours \\ \hline
BART-base & Instruction-tuned & 140 million & 3-5 hours \\ \hline
\end{tabular}%
}
\caption{Model related information (model type, number of parameters, training times).}
\label{tab:results-table-1}
\end{table*}

\begin{table*}[htbp]
\centering
\resizebox{0.6\textwidth}{!}{
\begin{tabular}{|l|l|l|l|l|}
\hline
\multicolumn{1}{|l|}{\textbf{Model}} & \multicolumn{1}{l|}{\makecell{\textbf{Accuracy}}} & \multicolumn{2}{l|}{\makecell{\textbf{Average inference}\\ \textbf{latency}}} & \multicolumn{1}{l|}{\makecell{\textbf{Model memory}\\ \textbf{footprint (MB)}}} \\ \hline
\multicolumn{1}{|l|}{} & \multicolumn{1}{l|}{} & \multicolumn{1}{l|}{\makecell{\textbf{per order}\\ \textbf{turn (ms)}}} & \multicolumn{1}{l|}{\makecell{\textbf{per}\\ \textbf{phoneme (ms)}}} & \multicolumn{1}{l|}{} \\ \hline
\multicolumn{1}{|l|}{Atlantis} & \multicolumn{1}{l|}{89.74\%} & \multicolumn{1}{l|}{\textbf{\textcolor{purple}{600}}} & \multicolumn{1}{l|}{\textbf{\textcolor{purple}{10.37}}} & \multicolumn{1}{l|}{\textbf{\textcolor{teal}{50 MB}}} \\ \hline
\multicolumn{1}{|l|}{Babylon} & \multicolumn{1}{l|}{\textbf{\textcolor{teal}{90.07\%}}} & \multicolumn{1}{l|}{83} & \multicolumn{1}{l|}{1.59} & \multicolumn{1}{l|}{118 MB} \\ \hline
\multicolumn{1}{|l|}{Camelot} & \multicolumn{1}{l|}{88.87\%} & \multicolumn{1}{l|}{\textbf{\textcolor{teal}{77}}} & \multicolumn{1}{l|}{\textbf{\textcolor{teal}{1.56}}} & \multicolumn{1}{l|}{96 MB} \\ \hline
\multicolumn{1}{|l|}{Delphi} & \multicolumn{1}{l|}{\textbf{\textcolor{purple}{88.84\%}}} & \multicolumn{1}{l|}{81.67} & \multicolumn{1}{l|}{1.58} & \multicolumn{1}{l|}{80 MB} \\ \hline
\multicolumn{5}{|l|}{\textit{\textbf{LLM based NMT models}}} \\ \hline
\multicolumn{1}{|l|}{Flan-T5} & \multicolumn{1}{l|}{N/A\textbf{*}} & \multicolumn{1}{l|}{\textcolor{purple}{\textbf{1319.40}}} & \multicolumn{1}{l|}{\textcolor{purple}{\textbf{25.53}}} & \multicolumn{1}{l|}{\textcolor{purple}{\textbf{280 MB}}} \\ \hline
\multicolumn{1}{|l|}{BART} & \multicolumn{1}{l|}{N/A\textbf{*}} & \multicolumn{1}{l|}{\textbf{\textcolor{purple}{1778.8}}} & \multicolumn{1}{l|}{\textbf{\textcolor{purple}{34.43}}} & \multicolumn{1}{l|}{\textbf{\textcolor{purple}{560 MB}}} \\ \hline
\end{tabular}%
}
\caption{Performances of all the model architectures against LLMs meant for Translation use cases across various metrics (a) accuracy (b) inference speed and (c) memory footprint. \textbf{*}Accuracy of BART and Flan-T5 is not reported as ASR errors (on word level) significantly cascade down to these models, see Section \ref{results} for explanation.}
\label{tab:results-table-2}
\end{table*}

\section{Experiments}
\subsection{Hardware Setup}
For training our models, we utilized AWS Cloud, specifically p3.16xlarge instances equipped with NVIDIA Tesla V100 GPUs. Each p3.16xlarge instance features Intel Xeon 2.3GHz CPUs, 488 GB of memory, and 8 NVIDIA Tesla V100 GPUs. We implement the Transformer-variants using PyTorch\footnote{https://pytorch.org/} v2.0.0. The LLM based models are from the Transformers library v4.0.1 \cite{wolf-etal-2020-transformers}. The training times, number of parameters and the sizes for each candidate model architecture are elaborated in Table \ref{tab:results-table-1}. Additional information related to training hyperparameters used are reported in Table \ref{table:parameters}.

\vspace{-5pt}
\subsection{Dataset}
\label{section:dataset}
For training, we used the synthetic data generation method proposed by \cite{arel2019conversational}, that simulates drive-thru customer orders. We used a total of ~250 million samples to train our models\footnote{Due to privacy and proprietary reasons, we are unable to reveal the name of the drive-thru restaurant as well as make the datasets and the code public.}. To evaluate the performance of the models, we use a test set containing 5587 samples/orders taken from the actual drive-thru locations. Our datasets include various accents (e.g., Southern), dialects (e.g., Texas), linguistic vernacular groups (eg., AAVE), and capture different dialog patterns and styles (e.g., short vs. long turns, faux duplicate turns as mentioned in Sec. \ref{related_work}).

\subsection{Model architectures and Training Hyperparameters}\label{model}

\paragraph{Atlantis (CNN + Bi-directional LSTM)}
A 1-D convolutional layer to encode input to a 2-layer bidirectional LSTM network, inspired by \cite{graves2012sequence}, trained with CTC loss \cite{graves2006connectionist}.

\paragraph{Delphi (LSTM encoder-decoder with Attention)}
3-layered LSTM encoder decoder architecture with 256-dimensional embeddings, following \cite{bahdanau2014neural} 

\paragraph{Camelot (Vanilla Transformer)} A base vanilla Transformer proposed by  \cite{vaswani2017attention} without any architectural upgrades.

\paragraph{Babylon (LSTM-Pooled Transformer)} A Bi-LSTM stack to encode the input sequence, followed by a pooling layer producing a 256 dimensional vector as input to the standard Transformer encoder-decoder. For more details, refer to Section \ref{babylon_architecture}. Training configurations for optimal results are in Appendix \ref{appendix:sec1}.

\subsubsection{LLM based translation models}
We compare the above baselines with two open source off-the-shelf LLM-NMT models: (i) \textit{flan-t5-small}\footnote{https://huggingface.co/google/flan-t5-small}  \cite{chung2024scaling} and (ii) \textit{bart-base}\footnote{https://huggingface.co/facebook/bart-base}\cite{DBLP:journals/corr/abs-1910-13461}

\subsection{Evaluation Metrics}
\textbf{Accuracy} This metric checks for a 1:1 match between the NLU model's output - sequence of transcodes - with the ground truth, and accounts for exact match of customer intents as well as menu entity names and number of the corresponding menu items.


\paragraph{Latency} Crucial for real-time applications such as our drive-thru use case, we benchmarked inference speed per phoneme and average latency across order turns of varying lengths to ensure smooth, delay-free interactions— enhancing customer experience.


\paragraph{Memory Footprint} We evaluate each model's memory footprint on the deployment hardware with smaller footprints being more desirable.


\paragraph{\textbf{BLEU} and \textbf{ROUGE}} We evaluated models using the widely recognized BLEU \cite{papineni-etal-2002-bleu} and ROUGE \cite{lin-2004-rouge} metrics but found them less informative for assessing customer intents represented as \textit{transcodes}. Therefore, we excluded them from the final evaluation and elaborate further on the findings in this regard in Appendix \ref{appendix:sec2}.

\section{Results} \label{results}
We summarize the results across the following metrics: (a) accuracy, (b) latency, and (c) memory footprint trade-offs and compare these results across the model architectures chosen earlier in Section \ref{model}.

\textbf{Babylon is the best-performing model in terms of accuracy.} From Table \ref{tab:results-table-2}, Babylon achieves the highest accuracy at 90.07\%. 
While Camelot has a slightly smaller memory footprint, with 88.87\% accuracy, we choose Babylon of the two since the 1.2\% lift in accuracy significantly affects the application's business metrics.

\textbf{Camelot has the fastest latency.} Camelot demonstrates the best average inference latency with 1.56 ms per phoneme and 77 ms per order turn, making it ideal for real-time applications. However, its lower accuracy compared to Babylon, particularly in handling sequential dependencies, makes it less suitable for our production needs (examples provided in \ref{appendix:sec3}). Moreover, increasing the number of parameters for Camelot did not improve the accuracy of the model significantly.


\textbf{Babylon manages memory footprint effectively.} While attention-based models (Babylon, Camelot and Delphi) have a larger footprint relative to Atlantis, Babylon remains manageable for edge hardware considerations with a memory footprint of 118 MB. LLMs such as BART are too cumbersome to deploy efficiently in our settings as opposed to other candidates.


\textbf{Flan-T5 and BART are not suitable for our use case.} There are a few major limitations for deploying these LLMs in production:

a) These models tend to hallucinate, generating repetitive or semantically similar yet invalid transcodes: BART often produces repetitive transcodes. Both Flan-T5 and BART generate inaccurate transcodes (resembling customer query than intent). Furthermore, the model's performance deteriorates with longer or more complex orders (> 3 intents). Flan-T5 is semantically more accurate and hallucinates less than BART overall; however, neither of them is performant for our use case.

b) Deployment of these models would not be feasible on edge hardware consisting of CPU compute due to their large memory footprint and high inference latency, rendering them unusable for real-time ordering. Refer (Table \ref{tab:results-table-2}.

c) BART and Flan-T5 operate more efficiently on word-level inputs but are not robust enough to noisy, error laden sub-word (phoneme) sequences coming from upstream ASR.

Finally, while this paper primarily focuses on English, it is important to note that our proposed approach generalizes well across other spoken languages due to its foundation in text-based representations such as International Phonetic Alphabet (IPA)\footnote{https://en.wikipedia.org/wiki/International\_Phonetic\_Alp
habet}.

\vspace{-5pt}
\section{Limitations and Future Work}
This work did not explore alternative pooling strategies such as CNN-based pooling that is typically used in vision-based transformers. Future work could evaluate whether this alternative would yield further improvements w.r.t model memory footprint. Additionally, this work currently uses greedy decoding strategy, and it remains to be seen if employing beam search decoding strategy would provide further accuracy gains. 

\section{Conclusion}
This work introduces Babylon, an edge-efficient architecture designed for NMT-based intent translation in drive-thru ordering systems. Babylon utilizes an LSTM-based token pooling mechanism alongside a standard Transformer to translate customer speech (phonemes) into actionable intents (which we call transcodes), enhancing the model's ability to manage multi-intent scenarios. The LSTM component provides a rolling memory for Transformer block, reduces the computational load of the Transformer, and mitigates the impact of noisy or erroneous ASR outputs. Our evaluation shows that Babylon offers a significant advantage in terms of accuracy, latency, and memory efficiency when compared to other NMT architectures (Flan-T5-small and BART), making it a robust choice for real-time edge deployment in noisy, outdoor SLU environments.

\bibliography{custom}

\appendix

\section{Appendix}
\label{sec:appendix}

\subsection{Training Hyperparameters} \label{appendix:sec1}
In Table \ref{table:parameters} below, we provide the hyperparameters we used to finetune the models used in this work. Note that, for all the LSTM based models, a dropout with a probability of 0.1 is applied to the LSTM layers to prevent overfitting, as suggested by \cite{Zaremba2014RecurrentNN}.




\subsection{Limitations of BLEU and ROUGE Metrics for Transcodes Evaluation} \label{appendix:sec2}
To gain better insight into why evaluation metrics such as BLEU and ROUGE are not applicable to our use case, we selected a random 200 orders and analyzed the output transcodes generated by our candidate model Babylon. We classified each output into one or more of the following eight categories of prominent errors:

\subsubsection{Lack of Semantic Understanding}
The primary limitation of using BLEU for our transcodes lies in their inability to accurately assess the semantic content of the outputs. These metrics focus on the exact matching of words and their order, which is not always indicative of the user intent captured within our transcodes. The following examples illustrate common scenarios where these metrics fall short in our use case:
\paragraph{Word Order Errors}
One significant limitation of the BLEU metric is its position independence. BLEU penalizes candidate transcodes that do not capture the exact order of the items, even if the semantics remain unchanged. For example:

\begin{flushleft}
\noindent\textit{Reference}: [1, \textcolor{orange}{fudge}, item, add, 1, \textcolor{blue}{cookie}, item, add]\\
\textit{Hypothesis}: [1, \textcolor{blue}{cookie}, item, add, 1, \textcolor{orange}{fudge}, item, add]\\
\textit{BLEU Score}: \textcolor{red}{0.759}\\
\end{flushleft}
Despite both sequences representing the same set of items, the BLEU score is penalized due to the difference in order. Conversely, consider below example:

\begin{flushleft}
\noindent\textit{Reference}: [1, \textcolor{orange}{fudge}, item, add, 1, \textcolor{blue}{cookie}, item, add]\\
\textit{Hypothesis}: [1, \textcolor{orange}{fudge}, item, add, 1, \textcolor{blue}{cookie}, item, add]\\
\textit{BLEU Score}: \textcolor{red}{1}\\

\end{flushleft}
The BLEU score is a perfect 1 when the order matches exactly, showing its insensitivity to semantically equivalent variations in order.

\paragraph{Synonyms} BLEU metric fail to account for synonyms that are common in our transcodes language.

\subsubsection{Severity of errors not considered}
Within a product use case, such as ours, agent's mistaking one item for another or the inability to delete / add / update items when the models should, can translate to significant operational and financial losses. Such critical errors cannot be adequately detected by evaluation metrics such as BLEU. For example:

\begin{flushleft}
\noindent\textit{Reference}: [1, \textcolor{orange}{chocolate fudge}, item, add, \textcolor{blue}{delete}]\
\noindent\textit{Hypothesis}: [1, \textcolor{orange}{chocolate fudge}, item, add]\\
\noindent\textit{BLEU Score}: \textcolor{red}{0.669}\\
\end{flushleft}

In this case, the item wasn't supposed to be added- perhaps leading to a bad customer experience in the drive-thru. However, BLEU scores it relatively high.

\subsubsection{Sentence length affects the scores}
n-gram based BLEU scores are unfairly affected by the length of the sentences. In drive-thru orders, which often contain long orders / sequences, the n-gram based BLEU score is particularly sensitive to mismatches. For instance:

\begin{flushleft}
\noindent\textit{Reference}: [yes, polar\_answer]\\
\noindent\textit{Hypothesis}: [yes, polar\_answer, \textcolor{blue}{thanks}]\\
\noindent\textit{BLEU Score}: \textcolor{red}{0.0}\\
\end{flushleft}

Here, the addition of \textit{thanks} in the hypothesis results in a BLEU score of 0, despite capturing the main intent (i.e., a yes) correctly.

\paragraph{Normalization errors}
BLEU metrics penalize based on exact matches, which can be problematic when the orders are not normalized (e.g., plurals vs. singular, inconsistent capitalization). For example:

\begin{flushleft}
\noindent\textit{Reference}: [2, \textcolor{blue}{c}one, item, add, 1]\\
\noindent\textit{Hypothesis}: [2, \textcolor{blue}{C}one, item, 'add, 1]\\
\noindent\textit{BLEU Score}: \textcolor{red}{0.286}\\
\end{flushleft}

Ideally, the BLEU score should be 1, but the scores were significantly lower despite the orders being identical.

\subsection{Examples to showcase performance of Camelot vs. Babylon} \label{appendix:sec3}

\paragraph{Camelot exhibits loss of sequential information.} \hspace{5pt}

\textbf{Input:} \textit{two scoop with sprinkles with medium waffle cone} 

\textbf{Camelot output:} [“new\_item”, "1", "scoop", "extras", "1", "sprinkles", "new\_item", "1", "Medium", "size", "1", "waffle cone", "quantity", "2", "add\_item"]

\textbf{Babylon output:}

[“new\_item”, "2 scoop", "extras", "1", "sprinkles", "new\_item", "1", "Medium", "size", "1", "waffle cone", "quantity", "1", "add\_item"]

The phrase \textit{"two scoop with sprinkles"} above involves a clear sequential relationship between \textit{"two"} and \textit{"scoop"}. Camelot misinterprets this by failing to preserve the sequence, leading to two separate items of one scoop each. This error happens because Transformers, while powerful at handling long dependencies, rely on positional encodings and can sometimes fail to capture nuanced sequential relationships—like associating the quantity "two" directly with \textit{"scoop"}.

\paragraph{Camelot shows inadequate local context representation.} \hspace{5pt}

\textbf{Input:} \textit{small caramel ice cream with whipped cream and ice cream cone}

\textbf{Camelot output:} ["new\_item", "1", "small", "size", "1", "caramel ice cream", "extras", "1", "whipped cream","new\_item", "1", "ice cream", "quantity", "1", "cone", "add\_item"]

\textbf{Babylon output:} ["new\_item", "1", "small", "size", "1", "caramel ice cream", "extras", "1", "whipped cream","add\_item", "new\_item", "1", "ice cream cone", "add\_item"] 

In the above case, the phrase "ice cream cone" is a compound noun that refers to a single item. Camelot incorrectly interprets it as two separate items ("ice cream" and "cone"), failing to capture the local relationship between the words. This happens because Transformers, with their global attention mechanism, often overlook the tight, local dependencies between nearby tokens, especially in phrases that represent singular entities (like "ice cream cone").

\paragraph{Babylon acts as an effective regularizer for erroneous phoneme sequences.} 
\vspace{5pt}
\textbf{Input (erroneous from upstream ASR):} \textit{small caremel ice cream with mcadam nuts}

\textbf{Correct Input:} \textit{small caramel ice cream with macadamia nuts}

\textbf{Camelot output:} ["new\_item", "1", "small", "size", "1", "caramel ice cream", "add\_item"]

\textbf{Babylon output:} ["new\_item", "1", "small", "size", "1", "caramel ice cream", "extras", "1", "macademia nuts", "add\_item"]

The error in the above case involves the substitution of the word \textit{"macadamia"} with \textit{"McAdam,"} a phonetically similar but incorrect entity for the domain. Camelot (vanilla Transformer) ultimately ignores this entity entirely. Babylon’s LSTM and token pooling architecture handles this error much better; it sequentially processes the phoneme input and retains a local context that helps identify familiar patterns like "macadamia nuts" with erroneous phonemes. Token pooling further regularizes the noisy phoneme output by focusing on key parts of the input and deemphasizing noisy ones.

\vspace{-10pt}

\onecolumn
\small
\begin{longtable}{|p{2.5cm}|p{1.5cm}|p{1.5cm}|p{1.5cm}|p{1.5cm}|p{1.5cm}|p{1.5cm}|}
\hline
\textbf{Parameter} & \textbf{Atlantis} & \textbf{Camelot} & \textbf{Babylon} & \textbf{Delphi} & \textbf{FlanT5} & \textbf{BART} \\ \hline
\endfirsthead
\multicolumn{7}{c}{{\bfseries \tablename\ \thetable{} -- continued from previous page}} \\
\hline
\textbf{Parameter} & \textbf{Atlantis} & \textbf{Camelot} & \textbf{Babylon} & \textbf{Delphi} & \textbf{FlanT5} & \textbf{BART} \\ \hline
\endhead
\hline \multicolumn{7}{|r|}{{Continued on next page}} \\ \hline
\endfoot
\hline
\endlastfoot
split\_test & 0.0 & 0.0 & 0.0 & 0.0 & 0 & 0 \\ \hline
split\_train & 100.0 & 100.0 & 100.0 & 100.0 & 100 & 100 \\ \hline
split\_validation & 0.0 & 0.0 & 0.0 & 0.0 & 0.0 & 0.0 \\ \hline
lr\_scheduler\_name & StepLR & StepLR & StepLR & StepLR & StepLR & StepLR \\ \hline
lr\_scheduler\_gamma & 0.1 & 0.1 & 0.1 & 0.1 & 0.1 & 0.1 \\ \hline
step\_interval & epoch & epoch & epoch & epoch & epoch & epoch \\ \hline
hidden\_size & 512 & 256 & 256 & 256 & 512 & 768 \\ \hline
optimizer\_name & RAdam & RAdam & AdamW & RAdam & AdamW & AdamW \\ \hline
optimizer\_lr & 1.0e-04 & 1.0e-06 & 1.0e-06 & 1.0e-04 & 6.4e-05 & 1.0e-06 \\ \hline
accumulate\_grad\_batches & 1 & N/A & 2 & N/A & 4 & 3 \\ \hline
batch\_size & 128 & 128 & 128 & 32 & 16 & 8 \\ \hline
clip\_grad & 0.05 & 0.05 & 0.05 & 0.2 & 0.05 & 0.05 \\ \hline
precision & float16 & float16 & float16 & float16 & float32 & float16 \\ \hline
\caption{Training hyperparameters used.}
\label{table:parameters}
\end{longtable}
\twocolumn

\end{document}